\documentclass[journal]{IEEEtran}
\usepackage[utf8]{inputenc}
\usepackage{times}
\usepackage{epsfig}
\usepackage{graphicx}
\usepackage{amsmath}
\usepackage{amssymb}
\usepackage{multirow}
\usepackage{cite}
\usepackage{booktabs}
\usepackage{balance}
\usepackage{color}
\usepackage{float}
\usepackage{pifont}
\usepackage[pagebackref=true,breaklinks=true,colorlinks,bookmarks=false]{hyperref}
\usepackage[T1]{fontenc}
\usepackage{siunitx}
\usepackage[dvipsnames,svgnames,x11names]{xcolor}
\newcommand{\cmark}{\ding{51}}%
\newcommand{\xmark}{\ding{55}}%

\newcommand{\etal}{\emph{et al.}}
\newcommand{\eg}{\emph{e.g.}}
\newcommand{\ie}{\emph{i.e.}}

\begin{document}
\title{ CLIP-SR: Collaborative Linguistic and Image Processing for Super-Resolution}
\author{Bingwen~Hu,
        Heng~Liu,
        Zhedong~Zheng,
        and Ping~Liu,~\IEEEmembership{Senior Member,~IEEE}
\thanks{Manuscript received April 19, 2005; revised August 26, 2015.}}%
\markboth{Journal of \LaTeX\ Class Files,~Vol.~14, No.~8, August~2015}%
{Shell \MakeLowercase{\textit{et al.}}: Bare Demo of IEEEtran.cls for IEEE Journals}

\maketitle

\begin{abstract}
Convolutional Neural Networks (CNNs) have significantly advanced Image Super-Resolution (SR), yet most CNN-based methods rely solely on pixel-based transformations, often leading to artifacts and blurring, particularly under severe downsampling rates (\eg, 8$\times$ or 16$\times$). The recently developed text-guided SR approaches leverage textual descriptions to enhance their detail restoration capabilities but frequently struggle with effectively performing alignment, resulting in semantic inconsistencies.
To address these challenges, we propose a multi-modal semantic enhancement framework that integrates textual semantics with visual features, effectively mitigating semantic mismatches and detail losses in highly degraded low-resolution (LR) images. Our method enables realistic, high-quality SR  to be performed at large upscaling factors, with a maximum scaling ratio of 16$\times$.
The framework integrates both text and image inputs using the prompt predictor, the Text-Image Fusion Block (TIFBlock), and the Iterative Refinement Module, leveraging Contrastive Language-Image Pretraining (CLIP) features to guide a progressive enhancement process with fine-grained alignment. This synergy produces high-resolution outputs with sharp textures and strong semantic coherence, even at substantial scaling factors.
Extensive comparative experiments and ablation studies validate the effectiveness of our approach. Furthermore, by leveraging textual semantics, our method offers a degree of super-resolution editability, allowing for controlled enhancements while preserving semantic consistency. The
code is available at \emph{https://github.com/hengliusky/CLIP-SR}.
\end{abstract}


\IEEEpeerreviewmaketitle

\section{Introduction}

The advent of Convolutional Neural Networks (CNNs) has significantly advanced the field of image super-resolution (SR) \cite{lim2017enhanced,wang2018esrgan,tian2020coarse, dong2015image, 9760261, 9318504}. Early CNN-based SR methods, which relied solely on low-resolution (LR) images to reconstruct high-resolution (HR) counterparts, often struggled to increase the reconstruction quality of their
outputs. To overcome these limitations, subsequent research \cite{chen2018fsrnet, shen2018deep, bulat2018super,guo2020learning, buhler2020deepsee, ma2020structure,li2023cross, zhao2024activating, liu2024cte} introduced prior information to guide the SR process, aiming to compensate for the missing details in LR images.
While prior-based approaches have demonstrated improvements, they tend to be restricted to specific types of images, such as those with well-defined structures or attributes (\eg, facial images). Moreover, methods such as SFTGAN \cite{wang2018recovering}, which leverage semantic segmentation maps to assist the SR reconstruction procedure, often introduce additional computational costs and are highly dependent on the accuracy of the segmentation process.

\begin{figure}[t]
	\centering
	\includegraphics{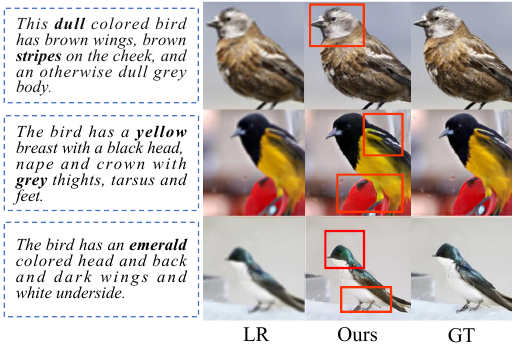}
    \vspace{-1em}
	\caption{Visualization of the results recovered by our method from low-resolution (LR) inputs. We highlight the semantic coherence part by aligning the textual guidance with the high-resolution (HR) ground truth.
 }
 \label{fig:1-1}
  \vspace{-1em}
\end{figure}

The use of text descriptions as a form of semantic guidance has
emerged as a more flexible and comprehensive alternative for addressing these limitations. Text offers richer and more detailed semantic information, which can guide the super-resolution process across a broader range of images. TGSR \cite{ma2022rethinking} was the first method to explore this approach; it uses text to enhance its ability to generate SR image details. However, challenges remain with regard to this method, particularly in terms of achieving effective text-image feature matching and semantic alignment, leading to inconsistencies between the input LR images and the generated SR results.
In this paper, we propose a novel approach that ensures semantic consistency while achieving large-scale super-resolution. Our method leverages text descriptions to guide the SR process, ensuring that the reconstructed HR images are both semantically coherent and visually realistic. As shown in Figure \ref{fig:1-1}, our approach addresses the limitations of the previously presented methods, providing a robust solution for conducting high-fidelity SR.

\begin{figure}[t]
\centering
\includegraphics{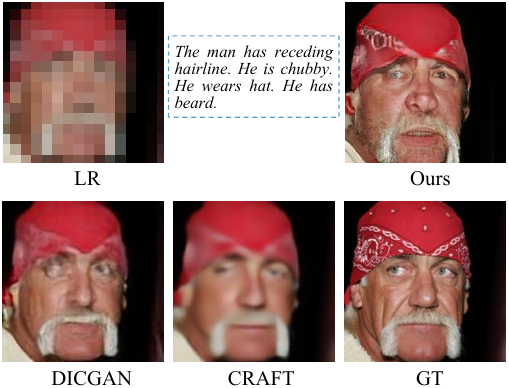}
\vspace{-1em}
\caption{An example comparison between the $16\times$ image SR results of our method and two SOTA SR methods: DICGAN \cite{ma2020deep} and CRAFT \cite{li2023feature}. Here, DICGAN and CRAFT are retrained on the same dataset as that used by our approach; LR is the input low-resolution image, and GT is the high-resolution ground truth (enlarged appropriately for visualization purposes).}
\label{fig:1-2}
\vspace{-1em}
\end{figure}

To address the challenges posed by the limitations of prior-based methods and ineffective text-image feature matching techniques, particularly when handling large-scale resolution degradation and high semantic ambiguity in real-world scenarios, we introduce a novel approach: Multi-modal Collaborative Semantic Enhancement for Super-Resolution (SR). Rather than treating the relevant text as mere prior guidance, we leverage the text information in conjunction with the LR image as two inputs modalities for SR tasks. Combining these modalities enhances their local semantics and enables high-performance, large-scale SR.
Specifically, we introduce a prompt predictor designed to extract essential semantic elements from the input text. Inspired by VPT~\cite{jia2022vpt} and GALIP~\cite{tao2023galip}, the prompt predictor incorporates a fully connected layer and a self-attention mechanism, serving as a text-driven attention module. Unlike directly inputting raw text vectors into the pre-trained CLIP-ViT, the refined text vectors generated by the prompt predictor enable CLIP-ViT to better align the semantic features between the text and image, thereby improving the quality of the produced cross-modal representation.

To further optimize text-image interactions, we introduce TIFBlock, a novel alignment and fusion module that is specifically designed to enhance the cross-modal integration process. Leveraging pre-trained models such as CLIP~\cite{radford2021learning} for the initial feature extraction step, TIFBlock effectively synthesizes and refines representations, resulting in a significant improvement in its text-to-image matching performance.
Building upon TIFBlock, we develop an iterative refinement module, which is a structure dedicated to iterative detail recovery and semantic enhancement. This module progressively refines local details, addresses blurred regions, and maintains semantic consistency across different iterations. A core component of the iterative refinement module is the inclusion of a customized residual connection that is tailored to our framework, which facilitates smooth feature propagation while preserving semantic integrity. The customized residual connection is seamlessly integrated within this module to further optimize the pixel transition and feature propagation tasks, ensuring robust multi-modal fusion. Together, these components align with our design objectives of delivering seamless and effective collaboration between modalities.

By integrating textual descriptions with LR images, the proposed method enhances SR by leveraging both linguistic semantics and visual features. The traditional SR methods rely solely on visual information and struggle to reconstruct fine details in severely degraded images. In contrast, our collaborative framework uses textual guidance to refine structures and textures, producing SR images that are both visually realistic and semantically aligned with the input text. As shown in Figure~\ref{fig:1-2}, our method achieves high-fidelity reconstruction effects for a $16\times$ downsampled facial image, demonstrating competitive performance with state-of-the-art SR techniques. Furthermore, it offers strong interpretability and ensures semantic consistency with the given text descriptions.

The primary contributions of this work are as follows:

\begin{itemize}
\item We introduce a new multi-modal semantic coherence approach for large-scale image super-resolution, generating semantically consistent and realistic high-resolution images from severely degraded low-resolution inputs.
\item We design a novel Text-Image Fusion Block (TIFBlock) and integrate it with a pre-trained cross-modality model to form an iterative collaborative fusion structure, enabling our framework to progressively restore image details while enhancing local semantics.
\item We investigate the impact of diverse textual semantics on image super-resolution. Comprehensive comparative experiments and ablation studies validate the effectiveness of our SR approach, which also maintains semantic coherence.
\end{itemize}

\vspace{-0.5em}
\section{Related Work}
\subsection{ Prior-Based Image Super-Resolution}
Single-image super-resolution (SISR) has become a dynamic area within end-to-end deep learning~\cite{yang2019deep}. 
The development of diverse models and mechanisms has significantly improved SR methods, particularly in terms of their pixel reconstruction and detail approximation capabilities.
Early SR approaches~\cite{dong2015image,kim2016accurate,kim2016,shi2016real,ledig2017photo,zhang2018image} usually assume that LR image pixels are obtained through bicubic downsampling performed on their HR counterparts.
These methods employ various deep mapping networks to directly reconstruct SR image pixels from LR inputs.
While these approaches can produce promising results on synthetic data with small-scale degradation, their effectiveness deteriorates significantly in real-world, large-scale degradation scenarios because of the full or partial loss of LR semantics.

To attain improved performance in real-world SR scenarios, numerous prior-based approaches, which deploy explicit or implicit priors to enrich the detail generation process, have been proposed.
A representative explicit method is reference-based SR~\cite{zheng2018crossnet,zhang2019image,yang2020learning,jiang2021robust}, which leverages one or more high-resolution reference images that share similar textures to those of the input low-resolution image to guide the process of generating an HR output. However, matching the features of the reference with a low-resolution input could be challenging, and these explicit priors may not be available. 

The recent methods, including FSRNet \cite{chen2018fsrnet}, DeepSEE~\cite{buhler2020deepsee}, and SFTGAN~\cite{shen2018deep}, have shifted toward leveraging implicit priors, yielding improved results by integrating prior information directly into the SR process. For example, FSRNet \cite{chen2018fsrnet} leverages geometric priors to improve the SR effects produced for facial images, whereas Zhang~\etal \cite{zhang2023multi} harnessed multi-view consistency. DeepSEE \cite{buhler2020deepsee} utilizes semantic maps to explore extreme image SR. SFTGAN \cite{shen2018deep} introduces image segmentation masks as prior features for facial image SR. Although they are effective, these implicit priors are often tailored to specific situations, such as restricted categories \cite{chan2021glean,pan2021exploiting} or facial images \cite{chen2018fsrnet, shen2018deep, wang2021towards,yang2021gan}, limiting their applicability to more complex, real-world SR tasks. Recent progress in single-image super-resolution has leveraged visual language models and text-guided techniques to achieve increased restoration quality. Methods such as TGSR~\cite{ma2022rethinking}, CoSeR~\cite{sun2024coser}, XPSR~\cite{qu2024xpsr}, and TGESR~\cite{gandikota2024text} incorporate text semantics as prior conditions, providing additional contextual guidance for the SR reconstruction procedure.

\vspace{-1em}
\subsection{Multi-modal Fusion Guided Image Generation}
Multi-modal fusion has become an increasingly prevalent approach in various visual tasks, such as image generation, style transfer, and image editing. For example, keypoints are commonly utilized in motion generation~\cite{suo2024jointly} and automatic makeup applications~\cite{huang2021real}. In text-based image synthesis scenarios, GAN-INT-CLS~\cite{li2020lightweight} employs text descriptions to generate images using conditional Generative Adversarial Networks (cGANs). To enhance the quality of image, Stack-GAN~\cite{zhang2017stackgan}, AttnGAN~\cite{xu2018attngan}, and DM-GAN~\cite{zhu2019dm} leverage multiple generators and discriminators. DF-GAN~\cite{tao2022df} simplifies the text-to-image synthesis process with a more streamlined and effective approach. LAFITE~\cite{zhou2022towards} introduces a contrastive loss based on the CLIP model~\cite{radford2021learning}, offering more accurate guidance for generating precise images.
In artistic style transfer cases, CLIPstyler~\cite{kwon2022clipstyler} enables domain-independent texture transfer from text descriptions to source images, whereas CLVA~\cite{fu2022language} employs a patchwise style discriminator to extract visual semantics from style instructions, thereby achieving detailed and localized artistic style transfer. SISGAN~\cite{dong2017semantic} pioneered the use of an encoder-decoder architecture for conducting text-based semantic editing on images. ManiGAN~\cite{li2020manigan} introduces a two-stage architecture with an attentional cropping module (ACM) and a deformable cropping module (DCM) to facilitate independent network training for text-based image editing. The lightweight GAN~\cite{li2020lightweight} further improves the efficiency of the process by applying a word-level discriminator. ManiTrans~\cite{wang2022manitrans} employs a pre-trained autoregressive transformer, utilizing the CLIP model~\cite{radford2021learning} for addressing semantic losses. More recently, Zeng~\etal~\cite{zeng2024instilling} developed a multiround image-editing framework using language guidance.

The emergence of large language models has further spurred advancements in the text-to-image generation field. DALL-E~\cite{ho2022cascaded} uses VQ-VAE~\cite{van2017neural} to decompose images into discrete tokens, framing image synthesis as a translation task. LDM~\cite{rombach2022high} applies diffusion models to latent image vectors, enabling an efficient training process with high-quality results. GLIDE~\cite{nichol2021glide}, which is a diffusion-based text-to-image generation model, uses guided diffusion to enhance the text-conditioned synthesis procedure. GALIP~\cite{tao2023galip} incorporates the CLIP model within adversarial learning for text-to-image synthesis purposes. ControlNet~\cite{zhang2023adding}, which is introduced by Zhang~\etal, builds upon the pre-trained Stable Diffusion~\cite{rombach2022high}, incorporating a detailed scheme control to guide the image generation process.

Recent advancements in pre-trained diffusion models~\cite{nichol2021glide, rombach2022high, saharia2022photorealistic} have significantly improved their image-generation capabilities. While studies \cite{hu2021lora, mou2024t2i, yue2024resshift, 10274147, yang2023towards} have underscored the generative potential of these models, applying them to SR remains challenging. The high fidelity required for SR demands both speed and efficiency—qualities that diffusion models generally lack due to their multi-step denoising process, which results in slower generation times and complicates latent space manipulation operations.

Compared with the use of pre-trained diffusion models, a GAN-based model is employed in this work for several key reasons. GANs facilitate high-resolution image generation in a single pass, which significantly improves upon the efficiency of diffusion models with an iterative nature. Furthermore, they provide a smooth latent space that enables intuitive control over the generated features, making them particularly well-suited for SR tasks. Additionally, GANs require less training data and computational resources, improving their accessibility for researchers. By leveraging GANs, we aim to achieve high-quality image generation while ensuring the practical applicability of super-resolution.

\vspace{-0.5em}
\section{Method}
In this section, we present an overview of our proposed CLIP-SR method, followed by detailed descriptions of each component contained within our multi-modal cooperative image super-resolution (SR) network. Finally, we introduce the total loss function used in our approach.

\begin{figure*}
\centering
	\includegraphics[width=0.95\textwidth]{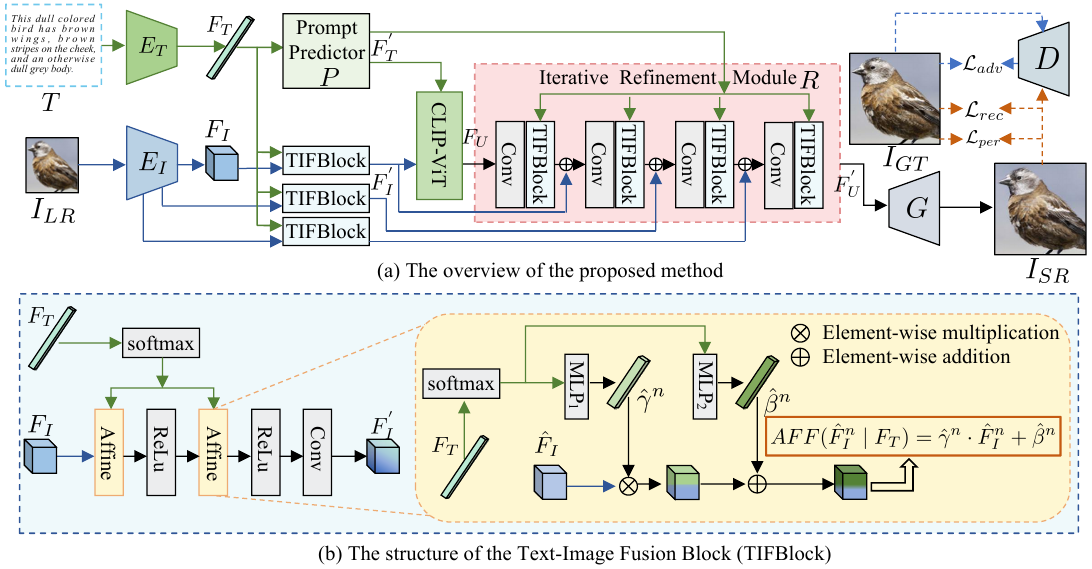}
    \vspace{-1em}
	\caption{  
 The architecture of our proposed multi-modal cooperative semantic enhancement model for large-factor image super-resolution (see subfigure (a)). Given an input low-resolution image $I_{LR}$ and text guidance $T$, features $F_{I}$ and $F_{T}$ are first extracted using an image encoder $E_{I}$ and a text encoder $E_{T}$, respectively. The text feature vector $F_{T}$ is further refined through a prompt predictor module $P$ and then processed by the CLIP-ViT model to enhance textual guidance. The refined text and image features are subsequently integrated using a Text-Image Fusion Block (TIFBlock), which aligns and combines the two modalities (see subfigure (b)). Within the TIFBlock, an affine transformation is applied in its text fusion module. This transformation employs two consecutive MLPs that generate channel-wise scaling parameters ($\hat{\gamma^{n}} = MLP_1(Softmax(F_{T}))$) and shifting parameters ($\hat{\beta^{n}} = MLP_2(Softmax(F_{T}))$). These parameters adaptively modulate the channel-wise features of the visual representation $\hat{F}_{I}^{n}$. Finally, the fused multi-modal features undergo iterative refinement and semantic enhancement through a continuous Conv-TIFBlock structure, which is referred to as an iterative refinement module $R$. This iterative process ensures that progressively improved super-resolution outputs with enhanced details and semantic coherence are obtained.}
\label{fig:3}
 \vspace{-1.em}
\end{figure*}

 \vspace{-1em}
\subsection{Overview}
The traditional small-factor SR methods generate HR images from LR images by using deep SR networks. However, large-factor downsampling operations often lead to significant blurring in LR images, making it challenging for SR networks to reconstruct semantically consistent and precise details solely from pixel-space information. To address these challenges, we introduce textual semantics as a complementary input, enabling our network to leverage information derived from both the pixel and textual spaces for generating more accurate details. 
For clarity, we denote the input low-resolution image as $L_{LR}$, the complementary text description as $T$, and the corresponding high-resolution ground truth as $I_{GT}$. The objective of CLIP-SR, denoted $\mathcal{H}$, is to fuse $L_{LR}$ and $T$ to generate a semantically consistent and visually realistic super-resolution image, which is denoted as $I_{SR}$.

Specifically, we introduce a text-image fusion block (TIFBlock) within a multi-modal iterative refinement model, which integrates CLIP \cite{radford2021learning} and a TIFBlock to effectively perform large-factor SR. To efficiently combine information derived from different modalities, \ie, text and images, we design a robust fusion strategy that preserves essential textual details while avoiding the information losses observed in simpler approaches~\cite{reed2016generative, dong2017semantic, zhang2017stackgan} that directly merge text vectors with image features. Our TIFBlock employs an affine transformation alignment strategy to increase the accuracy of text-to-image fusion and retain critical semantic details. Given the inherent differences between text and image features, precise alignment is crucial for achieving semantic coherence. To further reduce cross-modal inconsistencies, a prompt predictor is employed to process the text vectors prior to conducting alignment. Additionally, the CLIP model \cite{radford2021learning} is integrated within our framework as a supplementary alignment tool, ensuring a contextually precise and semantically coherent text-image fusion process for SR.
To ensure coherence with the LR content contained in the generated SR image, we design two additional mechanisms that build on our fusion strategy. Specifically, we incorporate residual connections to preserve the essential LR details, particularly in cases where semantic conflicts may arise. Additionally, text semantics are integrated at each layer of the multi-modal iterative refinement module, progressively guiding the SR process with fine-grained adjustments. 
These refined semantic fusion strategies ensure that the generated SR image remains both structurally and semantically consistent with the LR input. Figure \ref{fig:3} provides an overview of the overall network architecture and the details of the TIFBlock.

\vspace{-1em}
\subsection{Network Architecture}
In this section, we present the key components of our proposed multi-modal large-factor image super-resolution model. 
The model primarily comprises five components: text and image encoders, a prompt predictor, a text-image fusion block (TIFBlock), an iterative refinement module, and a CLIP-based discriminator. 

In essence, the text and image encoders extract text vectors and image features, respectively, providing foundational representations for the following steps. The TIFBlock aligns and fuses these features, enabling the cohesive integration of textual and visual information. CLIP-ViT and the prompt predictor effectively enhance the textual guidance provided throughout the generation process. The iterative refinement module progressively restores image details and enhances local semantics through multiple iterations, ensuring alignment between different modalities. 
Finally, the CLIP-based discriminator comprehensively evaluates the fidelity, semantic quality, and coherence of the generated image. By leveraging the synergistic interaction among these five components, our method generates semantically consistent and realistically reconstructed high-resolution images, even from severely degraded low-resolution inputs (\eg, with 8$\times$ or 16$\times$ downsampling).

\subsubsection{Text and Image Encoders}
We utilize two distinct encoders to process the input modalities. 
The text encoder, which is denoted as $E_{T}$, follows the architecture of CLIP \cite{radford2021learning} and encodes textual inputs $T$ into feature vectors $F_{T}$, where $F_{T} = E_{T}(T)$, to effectively capture semantic information.
For the input LR image $ I_{LR}$, the image encoder$E_{I}$ employs a series of convolutional layers to progressively transform the input into an $8\times8$ feature map $F_{I}$, where $F_{I} = E_{I}(I_{LR})$.
These encoders allow our model to generate compatible feature representations for both text and image inputs, preparing them for the subsequent fusion step within the network.

\subsubsection{Prompt Predictor}
Before leveraging the pre-trained CLIP-ViT model to align image features with corresponding text vectors, we introduce a prompt predictor inspired by VPT~\cite{jia2022vpt} and GALIP~\cite{tao2023galip}. The prompt predictor, which is denoted as $P$, comprises a fully connected (FC) layer and a self-attention layer; the predictor functions as a text-driven attention mechanism. It predicts text-conditioned prompts, $F_{T}^{'} = P(F_{T})$, which are appended to the visual patch embeddings in CLIP-ViT. This design enables the generated images to more effectively capture the semantic content of the input text while maintaining alignment with the visual information encoded by the CLIP-ViT model.

The prompt predictor leverages the output of the text encoder to selectively focus on salient textual elements, which are then fused with the visual features. 
This integration process enables the generator to more accurately interpret and translate the given text into detailed, coherent visual representations, enhancing the degree of alignment between the text descriptions and the generated images in terms of both content and quality.

\subsubsection{Text-Image Fusion Block (TIFBlock)}

To further enhance the influence of text information on images, we introduce a Text-Image Fusion Block (TIFBlock) that integrates textual semantics as a complementary feature source.
As shown in Figure \ref{fig:3} (b), the TIFBlock incorporates an affine transformation within its text fusion module. 
Following the design principles of DF-GAN \cite{tao2022df}, we introduce a ReLU layer after each affine layer to increase the diversity of text-fused images by introducing nonlinear relationships.
Additionally, to improve the ability of the model to comprehend text descriptions, we apply a Softmax function to re-weight the text features before passing them to the affine layer. 
This re-weighting strategy allows for a smoother and more reliable integration of the text and image domains. 

The process of the TIFBlock starts by feeding the LR image $I_{LR}$ into the image encoder network $E_{I}$, extracting an image feature vector $F_{I}$. Moreover, the text is encoded via the pre-trained CLIP encoder $E_{T}$, producing a text vector $F_{T}$. The text features are then re-weighted via the Softmax function before being passed through the affine transformation layer. 
Within this layer, the re-weighted text vector is processed through two consecutive Multi-Layer Perceptrons (MLPs), which generate a channel-wise scaling parameter $\hat{\gamma} = MLP_1(Softmax(F_{T}))$ and the channel-wise shifting parameter $\hat{\beta} = MLP_2(Softmax(F_{T}))$. 
The affine transformation then adaptively adjusts the channel-wise features of the visual feature $\hat{F}_{I}^{n}$. The affine transformation is defined as follows:
\begin{equation}
    AFF(\hat{F}_{I}^{n} \mid F_{T}) = \hat{\gamma}^{n} \cdot \hat{F}_{I}^{n} + \hat{\beta}^{n},
\end{equation}
where $AFF$ denotes the affine transformation, $\hat{F}_{I}^{n}$ represents the $n$-th channel of the visual feature map $\hat{F}_{I}$, $F_{T}$ represents the text vector, and $\gamma^{n}$ and $\beta^{n}$ are learnable scaling and shifting parameters, respectively. This mechanism enables the model to dynamically adjust the feature response to the textual context, leading to more accurate and meaningful alignment results.

The TIFBlock performs the initial alignment and integration steps on the text and image features by fusing these modalities through affine transformations, ensuring semantic consistency and accurate feature combinations. These fused multi-modal features are then passed to the Iterative Refinement Module, which progressively enhances the quality of the image by refining local details and reinforcing semantic coherence through multiple iterations. The iterative process builds on the fused features provided by the TIFBlock, enabling the model to generate outputs with higher resolution and realistic textures. Together, the TIFBlock establishes the foundational alignment of the two modalities, whereas the Iterative Refinement Module further optimizes and restores the image details in a step-by-step manner.

\subsubsection{Iterative Refinement Module} 
To ensure that the generated image aligns closely with the given text, we iteratively refine the image features derived from CLIP-ViT by using a residual structure to fuse text-image features in a process that is guided by the text vector. 
Initially, the prompt predictor leverages the output of the text encoder to bridge the semantic gap between the text and image modalities. 
The low-resolution image features $F_{I}$ are subsequently combined with the text vector $F_{T}$ within the TIFBlock to further align the image and text features. 
CLIP-ViT is then employed to reconcile any inconsistencies between the image and text, ensuring that the final image features match the knowledge existing in both modalities. 
Finally, the outputs acquired from the prompt predictor, TIFBlock, and CLIP-ViT model are iteratively merged via the residual structure to generate a high-resolution image that is semantically consistent with the provided text.

Throughout the entire pipeline, we utilize text information at three key stages. 
First, we employ a simple convolutional network to extract features from the low-resolution image, which are integrated with the text information using the TIFBlock. This integration scheme ensures that the combined features encapsulate both detailed visual cues and semantic information, enabling precise guidance for the information flow within the CLIP-ViT network. Next, a text attention mechanism processes the textual features to address the inherent differences between the text and image modalities, facilitating an effective cross-modal alignment process. Additionally, the textual information serves as the input of a prompt predictor that feeds into the CLIP-ViT model, further enhancing the fusion results obtained for visual and semantic features. Finally, after obtaining preliminary image features from CLIP-ViT, the iterative refinement module progressively restores detailed image information by iteratively fusing it with textual semantics and enlarging the image through an additional upsampling module $G$. The upsampling module $G$ consists of multiple blocks, each of which contains a 3$\times$3 convolutional layer (with a kernel size of 3, a stride of 1, and a padding of 1) followed by a PixelShuffle layer (with an upscaling factor of 2). The number of blocks is determined by the super-resolution scale factor.

\subsubsection{CLIP-Based Discriminator}
We utilize the CLIP-based discriminator proposed in GALIP \cite{tao2023galip}, which extracts more informative visual features from complex images, enabling the discriminator to more effectively identify unrealistic image regions. This, in turn, prompts the generator to produce more realistic images. The structure of the discriminator provides a deep understanding of complex scenes by integrating additional visual information into the CLIP framework, making it particularly well-suited for its role as a discriminator.
Specifically, the CLIP-based discriminator is designed to incorporate the language-image pre-training process of CLIP \cite{radford2021learning}, with enhancements tailored to improving its effectiveness at evaluating the quality of generated images.

During training, the discriminator aims to distinguish between generated and real images. The superior performance of the CLIP model in terms of aligning text and images derived from different modalities allows the CLIP-based discriminator to gain a comprehensive and nuanced understanding of the image content, contributing to the generation of higher-quality and semantically consistent outputs by our proposed method.

 \vspace{-1.em}
\subsection{Optimization Objectives}
\textbf{Reconstruction Loss.} To ensure consistency in the content of the reconstructed images, we employ the pixel-wise $\mathcal{L}_{1}$-norm, which is defined as follows:
\begin{equation}
\mathcal{L}_{rec}=\mathbb{E}[\|\mathcal{H}(I_{LR}, T) - I_{GT}\|_{1}],
\end{equation}
where $\mathcal{H}(I_{LR}, T)$ denotes the output generated by the full super-resolution network $\mathcal{H}$ proposed in this work, $F_{T}$ represents the text description, and $I_{GT}$ represents the high-resolution ground truth that corresponds to the input low-resolution image $I_{LR}$.

\textbf{Perceptual Loss.} Additionally, we use the perceptual loss \cite{johnson2016perceptual} to encourage visual consistency between the generated super-resolution results and the real high-resolution images. The perceptual loss is defined as follows:
\begin{align}
\label{eq5}
\begin{split}
    \mathcal{L}_{per} = \mathbb{E}\left[\sum_{i=0}^{5} \sigma_{i} \| \phi_{i}(\mathcal{H}(I_{LR}, T)) - \phi_{i}( I_{GT}) \|_{1}\right],
\end{split}
\end{align}
where $\phi_{i}(.)$ denotes the feature map derived from the $i$-th layer of the pre-trained perception network $\phi$. We employ the pre-trained VGG-19 network~\cite{Simonyan15} as our $\phi$ and select five activation layers for computing the perceptual loss. The hyper-parameters $\sigma_{i}$ modulate the contribution of the $i$-th layer to the total loss term in Equation~\ref{eq5}.

\textbf{Text-Constrained Adversarial Loss.} To constrain the semantic information contained in the input text, we utilize the text-constrained adversarial loss \cite{tao2023galip}. Here, $I_{LR}$ represents a given low-resolution image, and $F_{T}$ is the text vector extracted from the corresponding text input. Both the low-resolution image $I_{LR}$ and the text vector $F_{T}$ are fed into the super-resolution network $\mathcal{H}$, resulting in an output $\mathcal{H}(I_{LR}, T)$. 
Let $C$ and $\mathcal{V}$ represent the frozen CLIP-ViT model and the image feature extractor model contained in the CLIP-based discriminator, respectively. $Sim(.,.)$ denotes the cosine similarity between the generated HR image $\mathcal{H}(I_{LR}, F_{T})$ and the text vector $F_{T}$.

The text-constrained adversarial loss is defined as follows:
\begin{equation}
	\begin{aligned}
		\mathcal{L}_{adv} = & -\mathbb{E}_{(\mathcal{H}(I_{LR}, T) \sim \mathbb{P}_{g})} \left[ D(C(\mathcal{H}(I_{LR}, T), F_{T})) \right] \\
		& - \alpha \mathbb{E}_{(\mathcal{H}(I_{LR}, T) \sim \mathbb{P}_{g})} \left[ Sim (\mathcal{V}((\mathcal{H}(I_{LR}, T)), F_{T}) \right],
	\end{aligned}
\end{equation}
where $\alpha$ is a hyper-parameter that controls the weight of the text-image similarity, and $\mathbb{P}_{g}$ denotes the synthetic data distribution.

\textbf{Total Loss.} Considering all of the above loss functions, the total objective function is formulated as follows:
\begin{equation}
\mathcal{L}_{total} = \mathcal{L}_{rec} + \mathcal{L}_{per} + \lambda_{adv}\mathcal{L}_{adv},
\end{equation}
where the hyper-parameter $\lambda_{adv}$ controls the weight of the adversarial loss $\mathcal{L}_{adv}$.

\section{Experiment} 
\subsection{Implementation Details}
\textbf{Dataset.} 
We evaluate our method on the COCO~\cite{lin2014microsoft}, Caltech-UCSD Birds 200 (CUB)~\cite{WahCUB_200_2011}, and CelebA~\cite{cheng2015beyond} datasets, each of which contains images paired with textual descriptions. 
For training, all images are cropped to the resolution of $256\times256$, with low-resolution images generated by performing bicubic downsampling on high-resolution counterparts. The utilized CLIP-ViT backbone is the ViT-B/32 model.


\textbf{Setting.} We train the proposed method on an NVIDIA RTX A5000 by using the Adam optimizer with parameters of $\beta_1=0.0$ and $\beta_2=0.9$ over 220 epochs. The hyper-parameter $\lambda_{adv}$ is set to 0.01. Moreover, following the setup in GALIP \cite{tao2023galip}, we set $\alpha$ to 4. Since the official code for TGSR \cite{ma2022rethinking} is unavailable, we use TGSR${^{\#}}$ to represent results reproduced on the basis of the visual examples and quantitative metrics provided in the paper that propose TGSR for comparison with other methods.

\begin{figure*}[t]
\centering
	\includegraphics{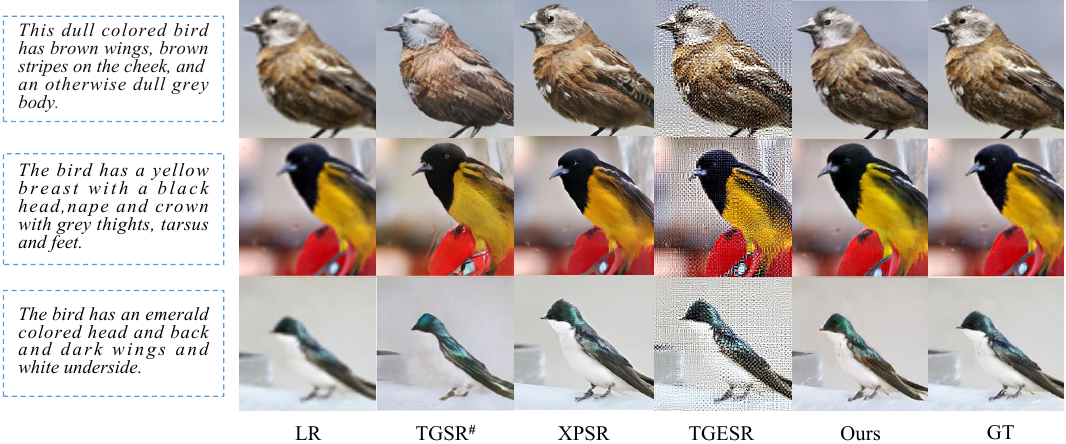}
    \vspace{-1em}
	\caption{Visual comparison among the $4\times$ SR results obtained with three SOTA SR methods, \emph{i.e.,} TGSR \cite{ma2022rethinking}, XPSR~\cite{qu2024xpsr}, and TGESR~\cite{gandikota2024text}, and our method on the CUB dataset. Notably, $\#$ indicates the results reported in the corresponding original paper. }
	\label{fig:4}
      \vspace{-0.5em}
\end{figure*}

\begin{table*}[t]
\centering
\vspace{-0.5em}
\caption{Quantitative comparison between our method and the comparative methods on the CUB and COCO datasets. The symbol $\downarrow$ denotes that lower values of the associated are better.} 
 \label{tab:2}
 \setlength{\tabcolsep}{12pt}{
		\begin{tabular}{l|c|cccccc}
		\toprule		Dataset&Metrics&Bicubic&EDSR\!\cite{lim2017enhanced}&ESRGAN\!\cite{wang2018esrgan}&SPSR\!\cite{ma2020structure} &TGSR${^{\#}}$\!\cite{ma2022rethinking} &Ours\cr
			\midrule
			\multirow{2}{*}{CUB}
			&NIQE~$\downarrow$ &12.374&10.684 &\textbf{5.465}&5.885&6.623&\underline{\bfseries5.825}\cr
			&PI~$\downarrow$ &9.747&8.168&\underline{\bfseries2.644}&3.345&\textbf{2.560}&4.167\cr
			\midrule
			\multirow{2}{*}{COCO}
	    &NIQE~$\downarrow$ &11.110&9.683&6.816&\underline{\bfseries6.378}&6.484&\textbf{4.706}\cr
		&PI~$\downarrow$ &9.373&8.515&7.135&6.060&\underline{\bfseries4.922}&\bfseries3.610\cr
			\bottomrule
		\end{tabular}
  }
  \vspace{-1em}
\end{table*}

\begin{table*}[t]
\centering
\vspace{-0.5em}
	\caption{Quantitative comparisons results obtained on the CelebA dataset.}
        \label{tab:3}
	\setlength{\tabcolsep}{10pt}{
		\begin{tabular}{l|cccccccc}
			\toprule
			Metrics&Bicubic&SuperFAN \cite{bulat2018super}&DICGAN \cite{ma2020deep}&TGSR${^{\#}}$ \cite{ma2022rethinking}&XPSR~\cite{qu2024xpsr} & TGESR~\cite{gandikota2024text}& Ours\cr
			\midrule  		
            PSNR~$\uparrow$&25.81&28.91&\bfseries33.61&23.48&26.76&  4.24 &\underline{\bfseries28.974}\cr	
            SSIM~$\uparrow$&\underline{\bfseries0.844}&0.815&\bfseries0.895&0.766&0.778 & 0.447 & 0.808\cr
                \midrule
        NIQE~$\downarrow$&14.514&6.459&\underline{\bfseries5.755}&8.846&6.511& 10.335&\bfseries5.172\cr	PI~$\downarrow$&9.676&5.345&5.599&7.165&\underline{\bfseries5.235}& 6.113 &\bfseries4.476\cr		
			\bottomrule
		\end{tabular}
	}	
       \vspace{-1em}
\end{table*}

 \vspace{-0.5em}
\subsection{Quantitative Evaluation}
To quantitatively assess the quality of the SR images generated by different methods, we utilize two primary evaluation metrics: the Natural Image Quality Evaluator (NIQE)\cite{wang2004image} and the Perceptual Index (PI)\cite{blau20182018}. The NIQE evaluates the overall quality of SR images, with lower scores indicating more natural and realistic results. The PI, on the other hand, measures the perceptual quality of the images, where lower PI values correspond to better visual quality. We specifically choose the NIQE and PI for our experiments (except for Table \ref{tab:3}) instead of traditional metrics such as PSNR and SSIM, which focus more on image distortion but overlook objective quality and perceptual experience. In the context of SR, the NIQE and PI are more aligned with assessing the realism and naturalness of images, making them better suited for this task.

Table \ref{tab:2} presents our experimental results obtained on the CUB and COCO datasets. For the smaller CUB dataset, we compare the NIQE and PI scores with those of several state-of-the-art super-resolution methods, including EDSR~\cite{lim2017enhanced}, ESRGAN~\cite{wang2018esrgan}, SPSR~\cite{ma2020structure}, and TGSR${^{\#}}$~\cite{ma2022rethinking}. Our method achieves the second-best NIQE score, closely following that of ESRGAN, while outperforming both Bicubic interpolation and EDSR in terms of the PI. On the larger COCO dataset, our approach significantly outperforms all the comparison methods in both the NIQE and the PI, demonstrating superior generalizability. The observed performance degradations exhibited by the other approaches on COCO further underscore the robustness and versatility of our method.

Table \ref{tab:3} provides quantitative comparisons among the PSNR, SSIM, NIQE, and PI metrics produced on the CelebA dataset. Our method is evaluated against several baseline approaches, including Bicubic interpolation, SuperFAN \cite{bulat2018super}, DICGAN \cite{ma2020deep}, TGSR${^{\#}}$ \cite{ma2022rethinking}, XPSR~\cite{qu2024xpsr}, and TGESR~\cite{gandikota2024text}. The results demonstrate that the proposed method achieves competitive performance across all the metrics. Specifically, compared with Bicubic interpolation, SuperFAN, and DICGAN, which rely solely on single-modality input, our approach incorporates supplementary textual information to achieve cross-modal semantic alignment, resulting in superior super-resolution performance. 
Moreover, in comparison with TGSR${^{\#}}$, XPSR~\cite{qu2024xpsr}, and TGESR~\cite{gandikota2024text}, which also utilize text guidance, our multi-modal collaborative semantic enhancement mechanism produces high-resolution images that are both semantically consistent and visually realistic.
In summary, our method consistently delivers competitive results across three datasets, underscoring its effectiveness in image super-resolution tasks.

\begin{figure*}[t]
\centering
	\includegraphics{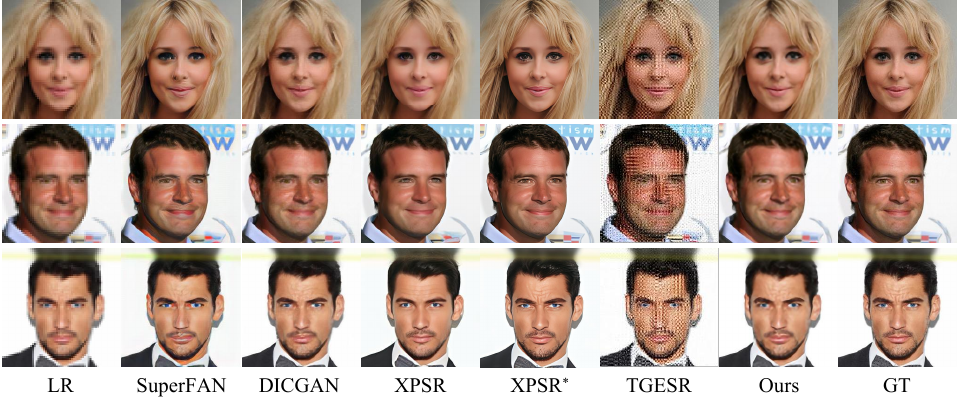}
    \vspace{-1em}
	\caption{Visual comparison among the $4\times$ SR results obtained with four SOTA SR methods, \emph{i.e.,} SuperFAN \cite{bulat2018super}, DICGAN \cite{ma2020deep}, XPSR~\cite{qu2024xpsr}, TGESR~\cite{gandikota2024text}, and our method on the CelebA dataset. $*$ denotes that 4$\times$ SR is applied on the basis of the settings of XPSR, where an input image with a 128$\times$128 resolution is upscaled to 512$\times$512.}
	\label{fig:5}
    \vspace{-1em}
\end{figure*}

\begin{figure*}[t]
\centering
	\includegraphics{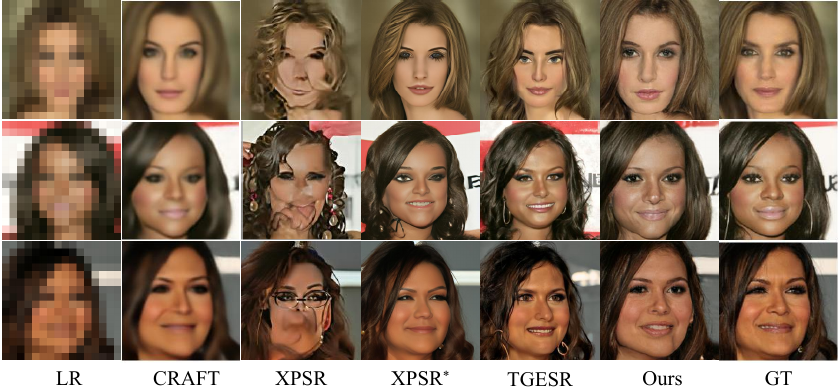}
    \vspace{-1em}
	\caption{Visual comparison among the $16\times$ SR results obtained with CRAFT \cite{li2023feature}, XPSR~\cite{qu2024xpsr}, TGESR~\cite{gandikota2024text}, and our method on the CelebA dataset. 
    $*$ indicates that 16$\times$ SR is applied on the basis of the settings of XPSR, where an input image with a 32$\times$32 resolution is upscaled to 512$\times$512.} 
	\label{fig:6}
\vspace{-1em}
\end{figure*}

\vspace{-1.0em}
\subsection{Qualitative Evaluation}
To further validate the effectiveness of the proposed method, we conduct additional qualitative experiments.
As illustrated in Figure~\ref{fig:4}, the experimental results demonstrate that our method achieves satisfactory visual outcomes even with this modification. These findings further confirm that the proposed multi-modal collaborative framework can consistently generate high-quality SR images with clear details and strong semantic coherence.

Concurrently, we conduct a $4\times$ SR experiment, upscaling low-resolution images from $64\times64$ to $256\times256$. As shown in Figure~\ref{fig:5}, our method, along with SuperFAN~\cite{bulat2018super}, DICGAN~\cite{ma2020deep}, XPSR~\cite{qu2024xpsr}, and TGESR~\cite{gandikota2024text}, achieves commendable visual quality. However, SuperFAN, DICGAN, and TGESR exhibit noticeable artifacts, whereas our approach produces smoother and more visually appealing results, highlighting its advantage with regard to leveraging text guidance for performing cross-modal semantic alignment.
Figure~\ref{fig:6} further presents the qualitative results of a $16\times$ SR task conducted on the CelebA dataset. CRAFT~\cite{li2023feature} generates overly smoothed images, failing to recover fine details. While TGESR produces visually plausible results, it struggles to preserve the semantic integrity of the source images. XPSR, though effective at $4\times$ SR, undergoes severe distortions at $16\times$ SR, even under its original experimental settings, demonstrating a substantial performance drop in large-scale SR cases with heavily degraded images.
In contrast, our method successfully super-resolves images to $256\times256$, achieving two key objectives: (1) restoring the essential semantic information and (2) maintaining high consistency with the original low-resolution input.

To compare the complexity and efficiency of our proposed SR model, we evaluate its number of parameters and inference time against those of several state-of-the-art models, including SuperFAN~\cite{bulat2018super}, XPSR~\cite{qu2024xpsr}, and TGESR~\cite{gandikota2024text}. The comparisons are conducted under the same conditions on an NVIDIA RTX 3090 GPU. As shown in Table~\ref{tab:4}, our method significantly outperforms the diffusion-based XPSR~\cite{qu2024xpsr} and TGESR~\cite{gandikota2024text} models in terms of model size and inference efficiency. However, owing to the incorporation of CLIP-ViT and the iterative refinement module, our model has a larger parameter count than SuperFAN~\cite{bulat2018super} does and has a longer inference time. Nevertheless, given the superior SR performance of our approach, this trade-off is acceptable.

\begin{table}[t]
\centering
\vspace{-0.5em}
\caption{Complexity and runtime efficiency comparisons among different methods. The Runtime represents the time consumed for inferring each image.}
\label{tab:4}
\setlength{\tabcolsep}{4pt}{
\begin{tabular}{l|cccccc}
\toprule
Metrics&  SuperFAN~\cite{bulat2018super} & XPSR~\cite{qu2024xpsr} & TGESR~\cite{gandikota2024text} & Ours \cr
\midrule  		
Parameters &  1.3 M& 1.9 B & 5.5 B &  632.2 M\cr	
Runtime   & \SI{4.1}{\milli\second}  & \SI{6.3}{\second} &  \SI{39.9}{\second}  & \SI{24.1}{\milli\second}   &  \cr	
\bottomrule
\end{tabular}
}	
\vspace{-1.5em}
\end{table}

\begin{figure*}[t]
\centering
	\includegraphics{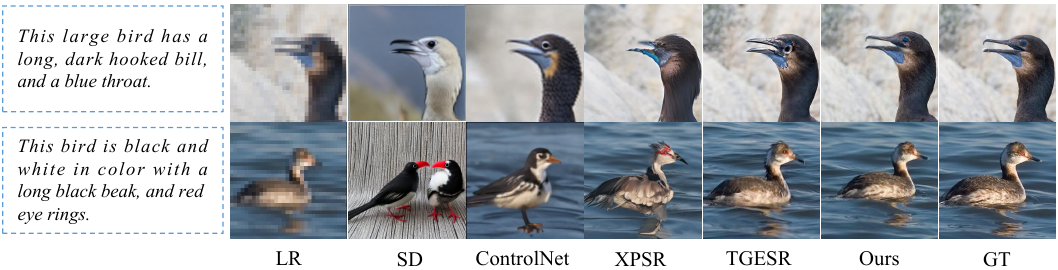}
    \vspace{-1em}
	\caption{Visual comparison among the $8\times$ SR results obtained with four diffusion model-based methods, \emph{i.e.,} Stable Diffusion (SD) \cite{rombach2022high}, ControlNet \cite{zhang2023adding}, XPSR~\cite{qu2024xpsr}, TGESR~\cite{gandikota2024text}, and our method on the CelebA dataset.}
	\label{fig:7}
    \vspace{-1em}
\end{figure*}

\vspace{-1em}
\subsection{Ablation Studies and Further Discussion}
To evaluate the effectiveness of each component included in our proposed method, we conducted ablation studies on the CUB dataset. We consider four variants: (1) a baseline U-Net for single image super-resolution, where $\mathcal{L}_{total}= \mathcal{L}_{rec} + \mathcal{L}_{per}$; (2) variant 1 with additional text supervision that incorporates our proposed multi-modal fusion architecture (including the TIFBlock and iterative refinement module), where $\mathcal{L}_{total}= \mathcal{L}_{rec} + \mathcal{L}_{per}$; (3) variant 2 with a pre-trained CLIP-ViT model, where $\mathcal{L}_{total}= \mathcal{L}_{rec} + \mathcal{L}_{per}$; and (4) variant 2 with a CLIP-based discriminator $D$,  where $\mathcal{L}_{total}= \mathcal{L}_{rec} + \mathcal{L}_{per} + \lambda_{adv}\mathcal{L}_{adv}$.

As shown in Table \ref{tab:5}, when the full model removes the text description module, CLIP-ViT, or the CLIP-based discriminator, the corresponding variants exhibit higher NIQE scores, indicating noticeable declines in performance. These experimental results demonstrate the effectiveness of incorporating textual information for enhancing the performance of the model. Additionally, they validate that the proposed Text-Image Fusion Block (TIFBlock) and Iterative Refinement Module effectively align textual and visual features, providing crucial semantic guidance for generating semantically consistent and realistic high-resolution images.

\begin{table}[t]
\centering
\vspace{-0.5em}
\caption{Comparison among the quantitative results obtained with different components of our method on the CUB dataset.}
\label{tab:5}
\setlength{\tabcolsep}{2.5pt}{
\begin{tabular}{l| c c c c | c c c c}
\toprule
Variants & U-Net & Text & CLIP-ViT & D & NIQE~$\downarrow$ & PI~$\downarrow$ &SSIM~$\uparrow$ & PSNR~$\uparrow$\\ 
\midrule
1 & \cmark & \xmark & \xmark & \xmark & 13.057 & 10.384 & 0.391 & 16.150\\
2 & \cmark & \cmark & \xmark & \xmark & 6.244 & 5.855& 0.834 & 26.987\\
3 & \cmark & \cmark & \cmark & \xmark & 6.578 & 6.020& 0.835 & 27.891\\
4 & \cmark & \cmark & \xmark & \cmark & 6.178 & 6.004 & 0.835 & 27.210\\
\midrule
Ours & \cmark & \cmark & \cmark & \cmark & \textbf{5.825} & \textbf{4.167} & \textbf{0.845} & \textbf{28.495}\\ 
\bottomrule
\end{tabular}
}
\vspace{-1em}
\end{table}

\vspace{-1em}
\subsection{Analysis of the Capability of LR-to-SR}
To evaluate the text-guided SR performance of our method, we conduct a comparison with Stable Diffusion (SD) \cite{rombach2022high}, ControlNet \cite{zhang2023adding}, XPSR~\cite{qu2024xpsr}, and TGESR~\cite{gandikota2024text} on the 8$\times$ SR task. These methods take low-resolution ($32\times32$) images along with textual descriptions as inputs. As shown in Figure~\ref{fig:7}, these models often introduce unwanted modifications, distorting the original visual information. In contrast, our method consistently produces sharper, more detailed SR images while effectively preserving both semantic coherence and fine-grained textures.

Our method outperforms SD and ControlNet in SR tasks for two main reasons. First, SD and ControlNet rely on iterative denoising, which struggles with performing 8$\times$ upscaling on severely degraded low-resolution images. While ControlNet introduces LR images as conditions, it lacks a mechanism for conducting semantic enhancement at extreme scaling factors. In contrast, our TIFBlocks iteratively refine both local textures and global semantics, producing more realistic and coherent SR results. Second, SD and ControlNet process image and text inputs separately, which can lead to misalignment between the textual descriptions and the generated images. Our approach employs a prompt predictor and iterative refinement module that leverage CLIP-based multi-modal alignment, ensuring semantic consistency and generating text-guided high-resolution outputs.

To further assess the effectiveness of text-guided SR methods, we conduct extensive experiments on multiple datasets, evaluating the tested methods, including XPSR and TGESR, across 4$ \times$, 8$ \times$, and 16$ \times$ SR tasks. As shown in Figure~\ref{fig:5}, XPSR achieves satisfactory results at 4$ \times$ SR for facial images but results in significant artifacts and structural distortions when it is applied to 8$ \times$ and 16$ \times$ SR tasks (see Figure~\ref{fig:6} and Figure~\ref{fig:7}, respectively), indicating its limitations in terms of handling extreme upscaling. Conversely, TGESR performs well at higher scaling factors (8$ \times$ and 16$ \times$ SR) but struggles to maintain fine-grained details at 4$ \times$ SR, suggesting an inconsistency in its ability across different upscaling levels. In comparison, our proposed CLIP-SR approach consistently generates high-quality super-resolved images across all scales, preserving both their semantic integrity and visual fidelity. As summarized in Table~\ref{tab:4}, CLIP-SR not only outperforms XPSR and TGESR in terms of reconstruction quality but also has superior computational efficiency and reduced model complexity. These results highlight the robustness and scalability of our approach, making it well-suited for high-fidelity SR applications implemented under varying degradation conditions.

\vspace{-1.0em}
\subsection{Analysis of the Editability of LR-to-SR Transformation}
\begin{figure}[t]
	\centering
	\includegraphics[width=\linewidth]{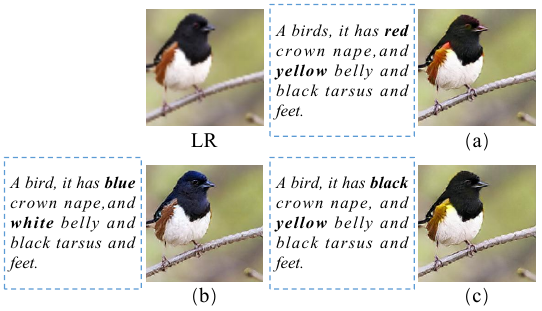}
    \vspace{-2.5em}
	\caption{Visualization of the results generated by our method under different text prompts. Our method demonstrates the ability to generate diverse and semantically consistent results.}
 \label{fig:8}
 \vspace{-1.0em}
\end{figure}

To evaluate the editability of our model in low-to-high-resolution transformations, we manipulate a subset of the CUB test images, as shown in Figure~\ref{fig:8}. Figure~\ref{fig:8}(b) presents color modifications in the nape, crown, and abdomen regions. Owing to low-resolution constraints, the network prioritizes pixel-level accuracy over high-level semantics, leading to slight blurring in the black region around the bird’s head. Nevertheless, our method successfully adjusts the wing color in the abdomen area. When prompted with “yellow” (Figure~\ref{fig:8}(c)), the model effectively alters the wing hue, exhibiting variations across the outputs. This diversity underscores its ability to perform semantically consistent, controllable edits.

\vspace{-0.5em}
\subsection{Analysis of the Effectiveness of Multi-modal Fusion and the Number of Iteration Layers}
\begin{figure}[t]
	\centering
	\includegraphics{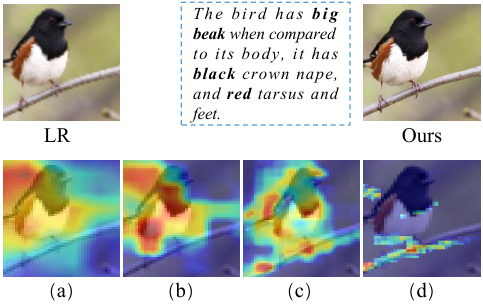}
    \vspace{-1em}
	\caption{Visualization of the heatmaps produced for low-resolution images during the super-resolution process. Subfigures (a), (b), (c), and (d) correspond to the results of performing text fusion at the first, second, third, and fourth layers, respectively, within the iterative refinement module}
  \label{fig:9}
  \vspace{-1.0em}
\end{figure}

To evaluate the effectiveness of our multi-modal fusion module and the impact of different numbers of iterative fusion layers, we analyze the heatmap outputs produced across different layers within the multi-modal fusion module. As shown in Figure~\ref{fig:9}, each text input is paired with a corresponding low-resolution image. Figure~\ref{fig:9} (a) shows the output derived from the initial text fusion layer, where the network begins by generating an image that is loosely aligned with the bird. In the subsequent layers, the attention of the model is progressively refined: in Figure~\ref{fig:9} (b), the focus shifts to the bird’s neck and body, whereas the further iterations shown in Figures~\ref{fig:9} (c) and (d) progressively enhance finer details, including the bird’s feet and tarsus. These findings empirically confirm the effectiveness of our iterative refinement module, demonstrating that four iterations are sufficient for achieving high-quality, semantically consistent text-to-image super-resolution results.

\begin{figure}[t]
	\centering
	\includegraphics{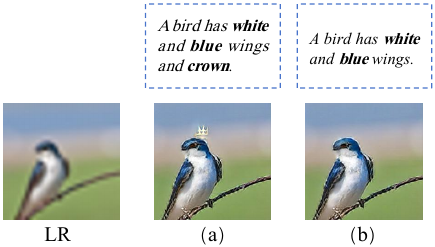}
    \vspace{-1em}
	\caption{Visualization of the results generated by our method under different text prompts. Given a low-resolution input image (a), our method produces super-resolution images guided by two distinct text descriptions: (b) and (c). While our method yields impressive results, certain details in the generated images may exhibit deviations due to the inherent ambiguity of natural language semantics.}
 \label{fig:10}
 \vspace{-1.0em}
\end{figure}
 
\vspace{-1em}
\subsection{Limitations}
Despite the superiority of the proposed method in the text-to-image super-resolution task, certain limitations warrant consideration in future research. The CLIP-ViT-B/32 model effectively leverages textual information to achieve enhanced image quality, particularly in the realm of semantic-guided super-resolution. It effectively bridges the gap between textual and visual data, enabling precise control over the process of generating high-resolution images. However, despite these strengths, the model can occasionally misinterpret ambiguous descriptions. For example, as illustrated in Figure~\ref{fig:10} (a), when instructed to generate an image featuring a "crown" on a bird, the model may incorrectly interpret the "crown" as a royal crown rather than the bird's crest. This misinterpretation underscores the necessity of including precise language in prompts. As demonstrated in Figure~\ref{fig:10} (b), removing the term "crown" and providing a more specific context often yields the desired image. Future research could focus on enhancing the ability of the model to disambiguate homonyms and develop a deeper understanding of context-specific semantics.

\vspace{-0.5em}
\section{Conclusion}
We introduce a multi-modal semantic consistency framework for large-scale image super-resolution (SR) that leverages text-image fusion to enhance both the visual fidelity and semantic coherence of images. Our approach integrates a pre-trained cross-modal model within an iterative refinement process, enabling progressive detail recovery and text-guided enhancements. Extensive experiments demonstrate that our model achieves superior performance to that of the existing methods, particularly in terms of preserving fine-grained textures and maintaining semantic alignment. Despite these advancements, our method struggles with ambiguous textual inputs, which can lead to inconsistencies in the SR results. Future work will focus on refining text preprocessing techniques to attain improved instruction clarity, further enhancing the controllability and reliability of text-guided SR.




\ifCLASSOPTIONcaptionsoff
  \newpage
\fi



%

{\footnotesize
\bibliographystyle{IEEEtran}
\bibliography{egbib}
}

%

\end{document}